\documentclass[runningheads]{llncs}
\usepackage{graphicx}
\usepackage{comment}
\usepackage{amsmath,amssymb} 
\usepackage{color}

\usepackage{epsfig}
\usepackage{graphicx}
\usepackage{amsmath}
\usepackage{amssymb}
\usepackage{mathtools}

\usepackage[pagebackref=true,breaklinks=true,letterpaper=true,colorlinks,bookmarks=false]{hyperref}
\usepackage[normalem]{ulem}

\usepackage{url}
\usepackage{multicol}
\usepackage{color}
\usepackage{algorithm,algorithmicx,algpseudocode}
\usepackage{amssymb}

\usepackage{xspace}
\usepackage{makecell}
\usepackage{arydshln}
\usepackage{caption}
\usepackage{subcaption}
\usepackage{graphicx}
\usepackage{multirow}
\usepackage{cleveref}
\usepackage{physics}

\algrenewcommand{\algorithmiccomment}[1]{\hfill$\blacktriangleright$ #1}

\newcommand{\rmm}{\text{DM}^2}
\newcommand{\eg}{\textit{e.g.}}
\newcommand{\ie}{\textit{i.e.}}
\newcommand{\etal}{\textit{et al.}}
\begin{document}

\pagestyle{headings}
\mainmatter
\def\ECCVSubNumber{21}  

\title{Diversified Mutual Learning for\\Deep Metric Learning} 


\makeatletter
\newcommand{\printfnsymbol}[1]{%
  \textsuperscript{\@fnsymbol{#1}}%
}
\makeatother

\titlerunning{Diversified Mutual Metric Learning}
%
\author{Wonpyo Park\thanks{Equal contribution}\inst{1}\and
Wonjae Kim\printfnsymbol{1}\inst{1}\and
Kihyun You\inst{2,3}\and 
Minsu Cho\inst{2}
}
\authorrunning{Wonpyo Park, Wonjae Kim, Kihyun You, and Minsu Cho}
%
\institute{
Kakao Corporation, Seongam, Korea
\and 
POSTECH, Pohang, Korea \\
\and
Kakao Enterprise, Seongam, Korea
}
\maketitle


\begin{abstract}

Mutual learning is an ensemble training strategy to improve generalization by transferring individual knowledge to each other while simultaneously training multiple models. In this work, we propose an effective mutual learning method for deep metric learning, called {\em Diversified Mutual Metric Learning}, which enhances embedding models with diversified mutual learning. 
We transfer relational knowledge for deep metric learning by leveraging three kinds of diversities in mutual learning: (1) {\em model diversity} from different initializations of models, (2) {\em temporal diversity} from different frequencies of parameter update, and (3) {\em view diversity} from different augmentations of inputs.
Our method is particularly adequate for \textit{inductive transfer learning} at the lack of large-scale data, where the embedding model is initialized with a pretrained model and then fine-tuned on a target dataset. Extensive experiments show that our method significantly improves individual models as well as their ensemble. Finally, the proposed method with a conventional triplet loss achieves the state-of-the-art performance of Recall@1 on standard datasets: 69.9 on CUB-200-2011 and 89.1 on CARS-196.


\end{abstract}


\section{Introduction} \label{sec:intro}

Mutual learning \cite{zhang2018deep} is an effective ensemble training strategy to improve generalization ability of learners.
In mutual learning, multiple models, \ie, \textit{cohort}, learn and teach each other simultaneously during the training time.  
For example, in the work of \cite{zhang2018deep}, two classifier models with different initializations are trained to predict similar output distributions given the same input data. Despite its effectiveness, previous methods of mutual learning~\cite{anil2018large,zhang2018deep} are limited to the task of classification.
In this work, we propose a mutual learning method for deep metric learning, which aims at learning deep embedding models that maps similar instances to nearby points on a manifold in the embedding space and dissimilar instances apart from each other. 

 
 In metric learning, due to small scales of training datasets, the embedding models are commonly initialized using an ImageNet-pretrained network such as Inception-BN \cite{ioffe2015batch} or ResNet-50 \cite{DBLP:journals/corr/HeZRS15}, and then fine-tuned on a target dataset. 
The common use of inductive transfer learning in metric learning limits the diversity of embedding models in mutual learning. 
Unlike the previous settings~\cite{anil2018large,zhang2018deep}, where models in the cohort are trained from scratch with diverse initializations,
these models with the same backbone network prevents mutual learning from distilling knowledge from each other during training.

To tackle the issue, we propose an effective mutual learning method for deep metric learning, called \textbf{D}iversified \textbf{M}utual \textbf{M}etric Learning ($\rmm$), which enhances the generalization ability of embedding models with diversified mutual learning.
We transfer the relational knowledge with pairwise distance between instances while diversifying the parameter update paths of models of the cohort with our three diversities:
(1) \textbf{model diversity} to leverage different initializations of models, (2) \textbf{temporal diversity} to diversify the frequency of parameter update, and (3) \textbf{view diversity} to exploits diverse inputs with different augmentations.
Note that the model diversity is naturally induced by training multiple models as in the work of \cite{zhang2018deep}, thus being the core of mutual learning per se. In this work, we introduce two additional diversities to mutual learning and also propose a mutual learning method using relation matrices for metric learning.  

Extensive experiments on standard image retrieval datasets \cite{WahCUB_200_2011,KrauseStarkDengFei-Fei_3DRR2013,Song2016DeepML} for deep metric learning show that the proposed method, $\rmm$, significantly improves performance of individual models as well as their ensemble, compared to conventional deep metric learning.
Moreover, $\rmm$ is also an effective regularizer that prevents severe overfitting on small training datasets in the latter part of the training.
The benefit of $\rmm$ monotonically grows as we increase the number of models in the cohort.
The proposed method combined with a conventional triplet loss achieves the state-of-the-art performance of Recall@1 on the standard datasets: 69.9 on CUB-200-2011 and 89.1 on CARS-196.


\section{Related Work} \label{sec:relwork}

Our work encompasses the studies of deep metric learning, knowledge transfer, and mutual learning.
In this section, we summarize each and explain their relations to our work.

\subsection{Deep Metric Learning} \label{sec:relwork_metric}

\subsubsection{Loss Function}

Devising a better loss function is one of the main challenges of the recent deep metric learning studies.
One family of the loss functions is pairwise distance-based loss which samples positive or negative pairs within a given mini-batch.
The objective of the losses is to let the distance of the positive pair to be small, while that of negative pair to be large.
Contrastive loss \cite{hadsell2006dimensionality} samples pairs of any two examples, while the Triplet loss \cite{DBLP:journals/corr/SchroffKP15} samples triplets of anchor, positive and negative examples.
In Triplet loss, the distance between the anchor and positive is trained to be smaller than the distance between anchor and negative by a certain margin.
Extended from the two losses, several methods \cite{yideep2014,Song2016DeepML,ustinova2016learning} are proposed to fully explore pairwise relation within a mini-batch.
Other than pairwise ones, losses using proxy \cite{movshovitz2017proxynca,zhaiclassification2019,qian2019softtriple} are also proposed.
Those methods assign single or multiple proxy vectors for each class, and distances are calculated with the proxies rather than individual embeddings. 
For our experiments, we adopt a simple pairwise distance-based loss: Triplet loss.

\subsubsection{Pair Sampling}
Sampling pairs within a mini-batch is crucial to ensure a stable convergence and a better performance for the pairwise distance-based losses.
In \cite{DBLP:journals/corr/SchroffKP15}, semi-hard negative examples are sampled rather than the hardest ones, which improves the performance of Triplet loss for very large face datasets.
Wu \etal \cite{wu2017sampling} propose the distance-weighted sampling considering the distribution of random points in a unit-hyper sphere.
Recently, Wang \etal \cite{wang2019multi} cast a sampling problem to a pair weighting problem based on a gradient analysis.
Other than the pair sampling within a mini-batch, Suh \etal \cite{suh2019stochastic} devised a batch sampling method which considers an inter-class relationship.
For our experiments, we used the distance-weighted sampling as the pair sampling method.

\subsubsection{Multiple Heads}
Similarly to the ensemble of totally separate models, multiple heads that share the backbone network for better embeddings is studied intensively in recent work.
The main focus is to diversify the output of each head for robust embedding.
For example, the importance of data examples is re-weighted differently for each head based on the gradients of other heads \cite{opitz_2018_pami} or difficulty of the examples \cite{yuan2017hard}.
In \cite{sanakoyeu2019divide}, a dataset is divided into multiple subsets and each head is trained for different subsets.
Kim \etal \cite{Kim_2018_ECCV} applied gating attention mechanism to heads, and the gates are trained to minimize their spatial overlapping regions.
In \cite{JACOB_2019_ICCV}, each head receives different high-order moments of the feature.
Although multiple heads approach can be adopted to further improve the performance of the deep metric learning model, it is beyond the scope of the proposed method; the models we used in the experiments are single-headed.

\subsection{Knowledge Transfer and Mutual Learning}

\subsubsection{Knowledge Transfer}

Knowledge distillation \cite{hinton2015distilling,romero2014fitnets,zagoruyko2016paying,huang2017like,liu2019knowledge} aims to transfer the knowledge from a teacher to a student model.
The typical setup is to train a student model to mimic the outputs of a teacher model. 
Traditional knowledge distillation \cite{hinton2015distilling} uses class probability as the knowledge to transfer.
However, the encoding models of embedding learning do not produce a certain class probability.
Therefore in deep metric learning, DarkRank \cite{chen2018darkrank} transfers similarity ranks between embeddings, and concurrently, relational knowledge distillation~\cite{park2019relational} and compact networks~\cite{yu2019learning} proposed direct transfer of pairwise distance between embeddings.
Multi-type knowledge \cite{liu2019knowledge} transfers graph structure lies in instance features and applied the method to the classification task.
We also set a pairwise relation matrix of embeddings within a mini-batch, as the knowledge to transfer.

\subsubsection{Mutual Learning}

The learning procedure of knowledge distillation requires a fully-trained teacher model.
Therefore, practitioners should endure tedious and time-consuming two-stage training pipeline: train a teacher model, and then distill the teacher to a student.
To overcome this shortcoming, deep mutual learning \cite{zhang2018deep} proposes a single stage training pipeline for knowledge distillation. 
The method minimizes the KL divergence between the class probability of multiple models on-the-fly in the training phase, and \cite{anil2018large} scaled the method up to hundreds of GPUs by adopting distributed setup.
In this online setting of knowledge distillation, there is no clear distinction between the teacher and student as in the classical setting of knowledge distillation, but the \textit{diversity} among the cohort should be guaranteed to make the knowledge be transferred.

Although the effectiveness of mutual learning is proven for classification tasks, all these schematics have not been applied in other tasks.
In this paper, $\rmm$ aims to provide adequate techniques to diversify the models of the cohort.
The diversification techniques are particularly important in deep metric learning where the effect of original \textit{model diversity} is not guaranteed due to the same initial weights of the pretrained models.

\begin{figure*}[t]
    \centering
    \includegraphics[width=\linewidth]{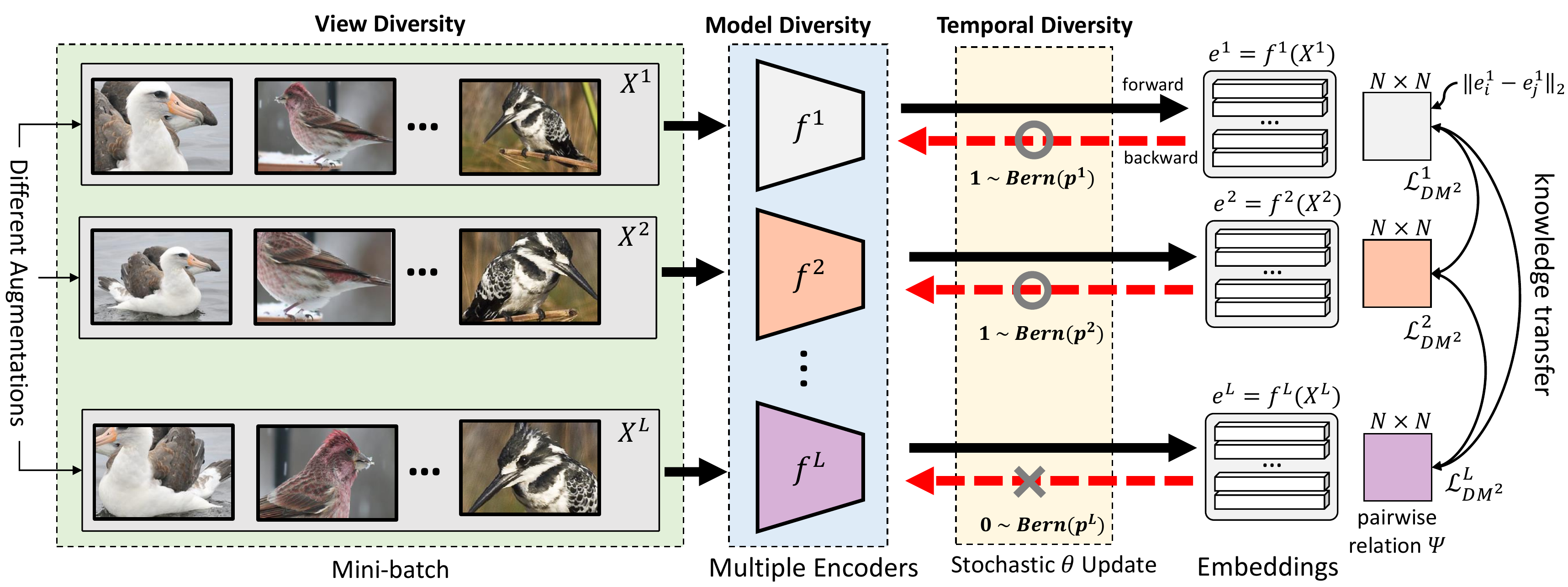}
    \caption{Given a mini-batch, $\rmm$ applies view diversity that exploits different augmentations for each encoder which are simultaneously trained. 
    To differentiate the frequency of parameter update among the cohort, $\rmm$ applies temporal diversity that stochastically updates the parameter of models according to their update probability. 
    The relational knowledge of each model is mutually shared between the models by transferring their pairwise distance of embedding vectors.}
\end{figure*}

\section{Diversified Mutual Metric Learning} \label{sec:method}

In the framework of the \textbf{D}iversified \textbf{M}utual \textbf{M}etric Learning ($\rmm$), multiple models, \ie, \textit{cohort}, are involved in the training process and transfer their knowledge to each other.
By diversifying the cohort, we have observed more diversity brings better generalization power for every single model and their ensemble.
In this section, we will define our objective function and deliver detailed explanations of our three diversities.

\subsubsection{Notation} Our cohort is comprised of multiple diverse models $f^i$, where $i$ is an index that goes from one to the size of the cohort $L$.
The model, or the encoder, $f^i$ maps an image $x$ into an embedding vector $e^i \in \mathbb{R}^d$, \ie, $e^i = f^i(x)$; for deep metric learning, the typical choice of $d$ is $128$ or $512$ \cite{wang2019multi,wu2017sampling}.
Then the distance between two images $x_1$ and $x_2$ can be measured via Euclidean distance of their respective embeddings: $\norm{f^i(x_1) - f^i(x_2)}_2 = \norm{e^i_1 - e^i_2}_2$.
Deep metric learning loss optimizes the encoders such that the distance between semantically similar images to be close and otherwise far apart.
Note that the architecture of the models can either be homogeneous or heterogeneous and its parameters can be diverse or identical, which we will treat as \textit{model diversity} in later \cref{sec:divs}.

\subsection{Objective Function}


We set the knowledge to be shared among the cohort to pairwise relation matrix of embeddings that each model produces for a given same mini-batch.
More concretely, for embeddings $\{e^i_{k}\}_{k=1}^{N}$ of model $f^i$, where $N$ is the size of mini-batch;
we define relation matrix of $f^i$ as $\Psi^i_{k,l} = \norm{e^i_k - e^i_l}_2$.
The knowledge sharing from $f_k$ to $f_l$ is defined as a difference between the relation matrix $\Psi$ as follows:

\begin{align} \label{eq:rkd-d}
     \mathcal{L}^{l\leftarrow k}_{\rmm} = \frac{1}{N^2} \sum_{i}^{N}\sum_{j}^{N} \norm{\Psi^l_{i, j} - \Psi^k_{i, j}}_2^2.
\end{align}

For each model $f^l$, the knowledge is transferred to all other models in the cohort $\{f^k\}_{k \neq l}$, defining the $\mathcal{L}_{\rmm}^l$ as follows:

\begin{align}
     \mathcal{L}^{l}_{\rmm} = \frac{1}{L-1}\sum_{k \neq l}^{L} \mathcal{L}^{l\leftarrow k}_{\rmm}.
\end{align}

The final objective $\mathcal{L}^{l}$ for the model $f^l$ in our diversified mutual metric learning is the sum of the deep metric learning loss $\mathcal{L}^l_{\text{DML}}$ and $\mathcal{L}^l_{\rmm}$ using Lagrangian multiplier $\lambda_{\rmm}$:

\begin{align}
    \mathcal{L}^{l} = \mathcal{L}^{l}_{\text{DML}} + \lambda_{\rmm}\cdot\mathcal{L}^{l}_{\rmm}.  
\end{align}

Throughout the experiments, we mostly used Triplet loss \cite{DBLP:journals/corr/SchroffKP15} for all $\{\mathcal{L}^l_{\text{DML}}\}_{l=1}^{L}$.
We applied a linear warm-up strategy for $\lambda_{\rmm}$, which makes each encoder to focus more on the metric learning loss during the early phase of the training.

\begin{center}
    \begin{algorithm}[t]
        \caption{Training procedure of $\rmm$}
        \begin{algorithmic}[1]
            \Require $L$ models $\{f^l\}_{l=1}^{L}$ with parameters $\{\theta^l\}_{l=1}^{L}$ \Comment{\textit{Model Diversity} (MD)}
            \For {the number of iterations}
            \State Sample a mini-batch $\{x_{k}\}_{k=1}^{N}$ of size $N$
                \For {$l$ = 1:$L$} \Comment{Done in parallel}
                    \State $\{A_{k}\}_{k=1}^{N} \sim \mathcal{A}$ \Comment{Sample the augmentations}
                    \State $\{x'_{k}\}_{k=1}^{N} = \{A_{k}(x_{k})\}_{k=1}^{N}$ \Comment{\textit{View Diversity} (VD)}
                    \State $\{e^l_{k}\}_{k=1}^{N} = \{f^l(x'_{k})\}_{k=1}^{N}$ \Comment{Infer embeddings}
                \EndFor
                \State (Barrier) Wait until the inference of entire cohort finished
                \For {$l$ = 1:$L$} \Comment{Done in parallel}
                    \State Compute relation matrix $\Psi^l$ from $\{e^l_{k}\}_{k=1}^{N}$ for $f^l$
                    \State Compute $\mathcal{L}^l_{\text{DML}}$ using $\{e^l_{k}\}_{k=1}^{N}$ for $f^l$
                    \State Compute $\mathcal{L}^l_{\rmm}$ using $\Psi^l$ and $\{ \Psi^k\}_{k \neq l}$ for $f^l$
                    \State $\mathcal{L}^{l} = \mathcal{L}^l_{\text{DML}} + \lambda_{\rmm}\cdot\mathcal{L}^l_{\rmm}$
                    \State$p \sim Bern(p^l)$ \Comment{Sample $p$ from a Bernoulli distribution} 
                    \If {$p$ is 1} \Comment{\textit{Temporal Diversity} (TD)}
                        \State Update $\theta^l$ w.r.t $\mathcal{L}^{l}$
                    \EndIf
                \EndFor
            \EndFor
        \end{algorithmic}
        \label{alg:rks}
    \end{algorithm}
\end{center}

\subsection{Diversities} \label{sec:divs}

We now deliver detailed explanations of three diversities that could make the cohort to share more rich knowledge for mutual learning.
The combination of the three diversities aims to diversify the update paths of multiple models in the parameter space, and the diversified paths lead to different local optima for each model. 
Due to the diversities, $\rmm$ can form a cohort of diverse models.
We concretely depict the general framework of $\rmm$ in \cref{alg:rks}.

\subsubsection{Model Diversity} The first is a \textit{model diversity} (MD), which diversifies the cohort by leveraging different initializations of models to encourage the individual models to explore a different region and settle at different local optima after training.
MD is naturally induced by the learning procedure of mutual learning \cite{zhang2018deep} where multiple models mutually teach each other.
However, this diversity is hard to be achieved in the setting of deep metric learning because models are initialized with the same pretrained parameters, which is essential in inductive transfer learning, do not allow much diverse models, and even forcing them to be identical.


\subsubsection{Temporal Diversity} The second is a \textit{temporal diversity} (TD), which diversifies the cohort by discriminating their frequency of parameter update. 
To discriminate the frequency of parameter update, we let the models stochastically update their parameters according to their update probabilities.
We assign diverse update probabilities $\{p^l = 2^{-(l-1)}\}_{l=1}^{L}$ to each model according to its index.
The first model has the highest update probability which is one, and the last model has the lowest update probability.
When the update probability is high, the model moves fast along the update path and it results in a more fitted model on the training dataset. On the contrary, when the update probability is low, the model results in less fitted on the dataset. Combining the models with different temporal steps in update helps to form a more diversified cohort.

\subsubsection{View Diversity} The third is a \textit{view diversity} (VD), which diversifies the cohort by exploiting diverse inputs with different augmentations for each model. 
Since the gradients of each model are computed using input data and activations, different augmentations to the mini-batch of each model yield diversified gradients. In addition, VD helps to learn a more robust embedding model, as the models learn to construct the relational structure $\Psi$ that is invariant to noise-injected data.

\subsubsection{} We defer the implementation details regarding the parameter initialization of the cohort ($\{\theta^l\}_{l=1}^{L}$) and the augmentation family $\mathcal{A}$ to \cref{sec:impl_detail}.

\subsection{Distributed parallel learning}
Computing $\mathcal{L}^l_{\rmm}$ requires only the relation matrices from the other encoders $\{ \Psi^k\}_{k \neq l}$.
Thus the computation including generating mini-batches with different views, inferring the encoder, and calculating the gradients can be done in parallel by allocating encoders to separate computational resources, \eg, GPU.
As a result, the training time of $\rmm$ almost stayed the same compared to that of a single model at the cost of computational resources.
\section{Experiment} \label{sec:experiment}

In this section, we report the efficacy of the $\rmm$, and the implementation details used in our experiments.
We conduct experiments on the following standard datasets for metric learning, and follow the evaluation protocol, and train/test splits as proposed in \cite{Song2016DeepML}.

\subsection{Dataset}

We run experiments on four datasets: CUB-200-2011 \cite{WahCUB_200_2011}, Cars-196 \cite{KrauseStarkDengFei-Fei_3DRR2013}, Stanford Online Products (SOP) \cite{Song2016DeepML} and In-Shop Clothes Retrieval (In-Shop) \cite{liu2016deepfashion}.

CUB-200-2011 consists of images of 200 bird species.
Half of 200 classes (5,864 images) are used for the training and the remaining 100 classes (5,924 images) are used for the testing.
Cars-196 consists of images of 196 car models.
As in the CUB-200-2011, half of 196 classes (8,052 images) are used for the training and the remaining 98 classes (8,131 images) are used for the testing.

SOP, which consists of images containing 22,634 product classes, is a much larger dataset compared to the former two datasets.
For the training, 11,318 classes (59,551 images) are used and the remaining 11,316 classes (60,499 images) are used for the testing.
In-Shop consists of images of 11,735 clothes products.
For the training, 3,997 classes (25,882 images) are used, and the remaining 7970 classes (26,830 images) are split into two subsets (query set and gallery set) for the testing.


\subsection{Implementation Detail} \label{sec:impl_detail}

Following the standard practice of deep metric learning studies, we use the inception network with the batch normalization (BN-Inception) \cite{ioffe2015batch} and the ResNet-50 \cite{DBLP:journals/corr/HeZRS15} pretrained on ILSVRC 2012-CLS (ImageNet) \cite{ILSVRC15} as our backbone networks.
We note again that these pretrained backbones make the effect of model diversity weaker.
However at the same time, the pretraining is strongly required for deep metric learning since the size of the datasets is very small to learn a visual feature extraction in an end-to-end manner, which makes the use of ImageNet pretraining essential.

We add a linear projection layer after the last pooling layer of a backbone network and apply $l_2$ normalization on the output embedding $e$ of the linear projection layer, \eg, $\frac{e}{\norm{e}_2}$.
We follow the image pre-processing and the augmentation family of a recent state-of-the-art method, HORDE \cite{JACOB_2019_ICCV}, which uses $256 \times 256$ random crop from randomly resized images and apply a random horizontal flip.

We train the parameters of batch normalization layers \cite{ioffe2015batch} in backbone networks for SOP and In-Shop datasets and freeze them for CUB-200-2011 and Cars-196 datasets following the procedure done in \cite{wang2019multi,qian2019softtriple}.

Regardless of the model architecture and dataset choice, the training was done with Adam optimizer \cite{kingma2014adam} with the initial learning rate of $3\cdot10^{-5}$ and weight decay of $5\cdot10^{-4}$; and the mini-batch size was $120$.
All experiments were conducted using pytorch 1.3.1 \cite{paszke2017automatic} with CUDA 10.2 on Nvidia V100 GPU. 

To make enough positive pairs within a mini-batch, we follow the batch construction of FaceNet \cite{DBLP:journals/corr/SchroffKP15}.
For every mini-batch, we randomly sample a certain number of classes and sample 5 images per class for small datasets (CUB-200-2011, Cars-196) and 2 images per class for large datasets (SOP, In-Shop).
We apply a warm-up strategy for $\lambda_{\rmm}$: linearly increasing it from $0$ to $20$ for the first three epochs of the training.

Unless specified, we adopt Triplet loss \cite{DBLP:journals/corr/SchroffKP15} as our deep metric learning loss $\mathcal{L}_{\text{DML}}$ and the used distance weighted sampling \cite{wu2017sampling} as a pair sampling method.
Note that we carefully tuned our hyper-parameters to build strong baseline models.
The source codes will be released in public upon acceptance.

\subsection{Efficacy of Diversities} \label{sec:div_eff}

\begin{table*}[t]
    \centering
    \setlength{\tabcolsep}{2pt}
    \caption{
    Recall@1 performance on CUB-200-2011 and Cars-196 datasets.
    We used cohort size of 4 for $\rmm$, and report how the performance increases with the added diversities: model diversity (MD), temporal diversity (TD), and view diversity (VD).
    We highlight the best performing single model and ensembled model. 
    \\
    }
    \label{table:efficacy_rmm}
    \begin{tabular}{|l|ccccc|ccccc|}
    \hline
    \multirow{2}{*}{Methods} & \multicolumn{5}{c|}{CUB-200-2011} & \multicolumn{5}{c|}{Cars-196} \\
    & Net A & Net B & Net C & Net D & Ens & Net A & Net B & Net C & Net D & Ens \\
    \hline
    \hline
    Independent & 66.07 & 66.18 & 67.21 & 65.97 & 70.71 & 85.24 & 84.78 & 85.22 & 85.26 & 89.87 \\
    \hline
    \hline
    MD & 67.93 & 67.12 & 67.18 & 67.32 & 69.60 & 85.79& 86.07 & 86.31 & 86.29& 87.86\\
    MD+TD & 69.45 & 69.01 & 68.10 & 67.59 & 71.25 & 89.00 & 88.49 & 87.54 & 86.21 & 90.49 \\
    MD+VD & 68.26 & 68.72 & 68.42 & 68.45 & 71.69 & 87.91 & 87.22 & 87.39 & 87.49 & 90.10 \\
    MD+TD+VD & \textbf{69.90} & 69.50 & 68.86 & 69.11 & \textbf{72.89} & \textbf{89.10} & 88.29 & 87.65 & 86.33 & \textbf{91.48} \\
    \hline
    \end{tabular}
\end{table*}

The result of \cref{table:efficacy_rmm} shows the effect of the diversities in the application of mutual learning for deep metric learning.
We used a cohort of size 4 for these experiments, the size of the embedding vector is set to 512, and the networks are all BN-Inception architecture initialized with the same parameter pretrained on ImageNet.
The only differences in the parameter are induced by the random initialization of the linear projection layer.

The first row of \cref{table:efficacy_rmm} (independent) indicates the results of four independently trained networks (\ie, $\lambda_{\rmm} = 0$) and their concatenation-based ensemble performance.
The MD only results are identical to the naive adoption of mutual learning.
As expected, the naive adoption brings worse result than the individual learning without construct cohort.
On the contrary, one can see that the addition of our proposed novel diversities TD, VD brings superior results; and applying all diversities brings the state-of-the-art single model performance (see \cref{tab:sota_metriclearning}).

\begin{figure}
    \centering
    \begin{subfigure}{0.47\linewidth}
        \label{fig:scale_cub200}
        \includegraphics[width=\textwidth,height=4cm]{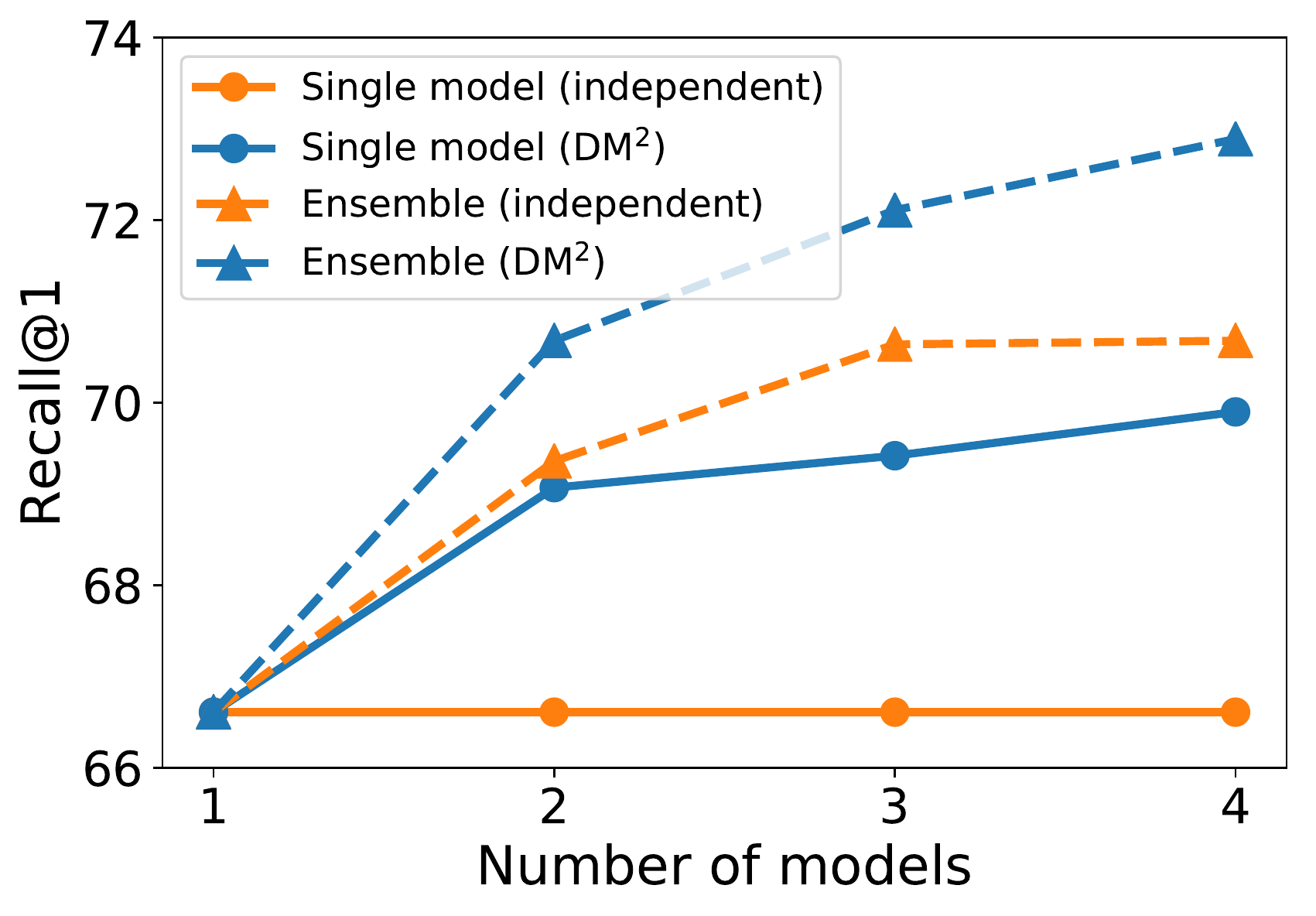}
        \caption{Test Recall@1 on CUB-200-2011}
    \end{subfigure}
    \begin{subfigure}{0.47\linewidth}
        \label{fig:scale_cars196}
        \includegraphics[width=\textwidth,height=4cm]{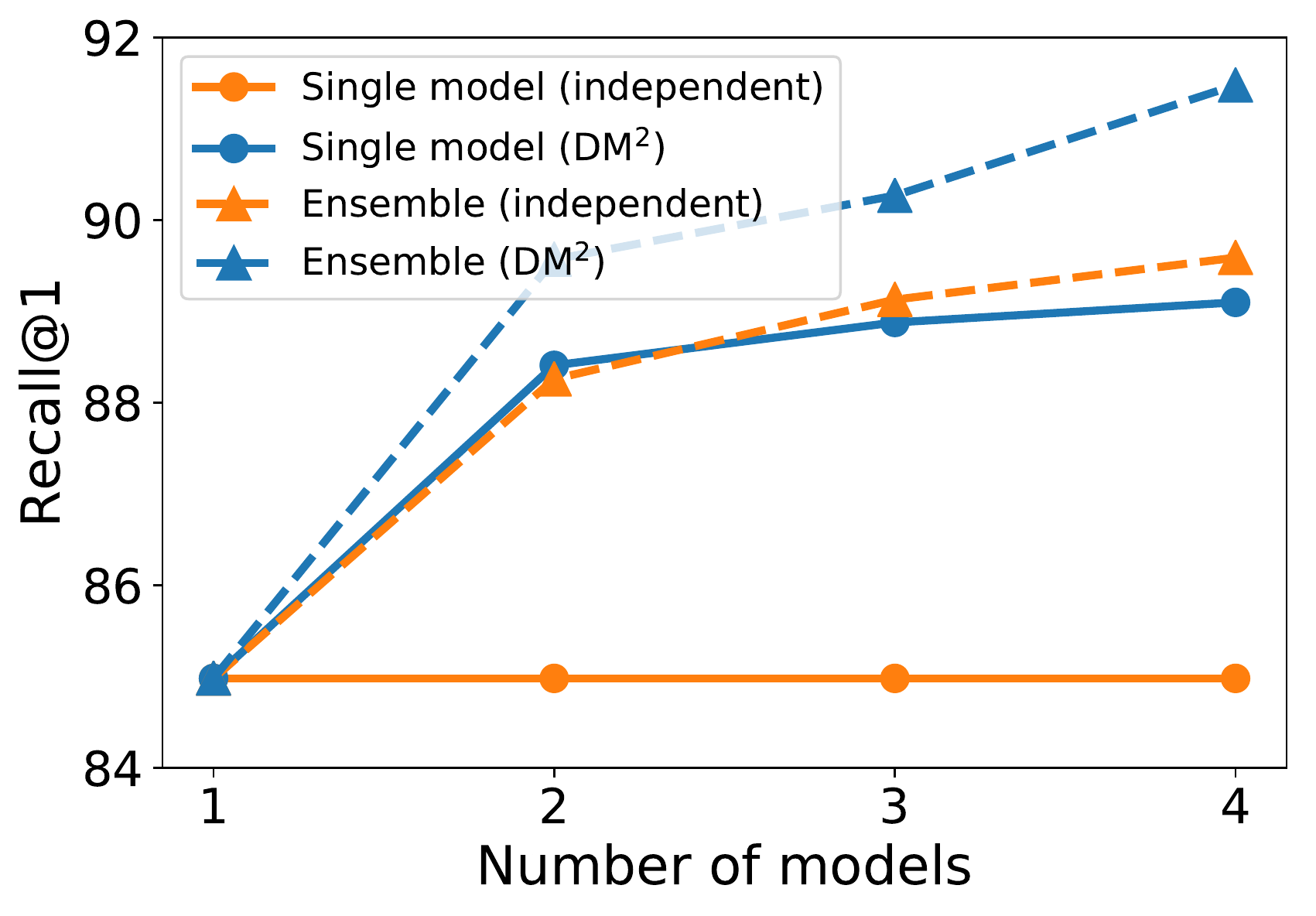}
        \caption{Test Recall@1 on Cars-196}
    \end{subfigure}
    \caption{Plots showing Recall@1 on the evaluation sets of CUB-200-2011 and Cars-196 with various cohort size ($1$, $2$, $3$, $4$). 
    For the single model results of $\rmm$, we report the result of the first model ($l$=$1$) among the multiple models of the cohort.}
    \label{fig:scale}
\end{figure}

\begin{figure}
    \centering
    \begin{subfigure}{0.47\linewidth}
        \label{fig:reg_cub200}
        \includegraphics[width=\textwidth,height=4cm]{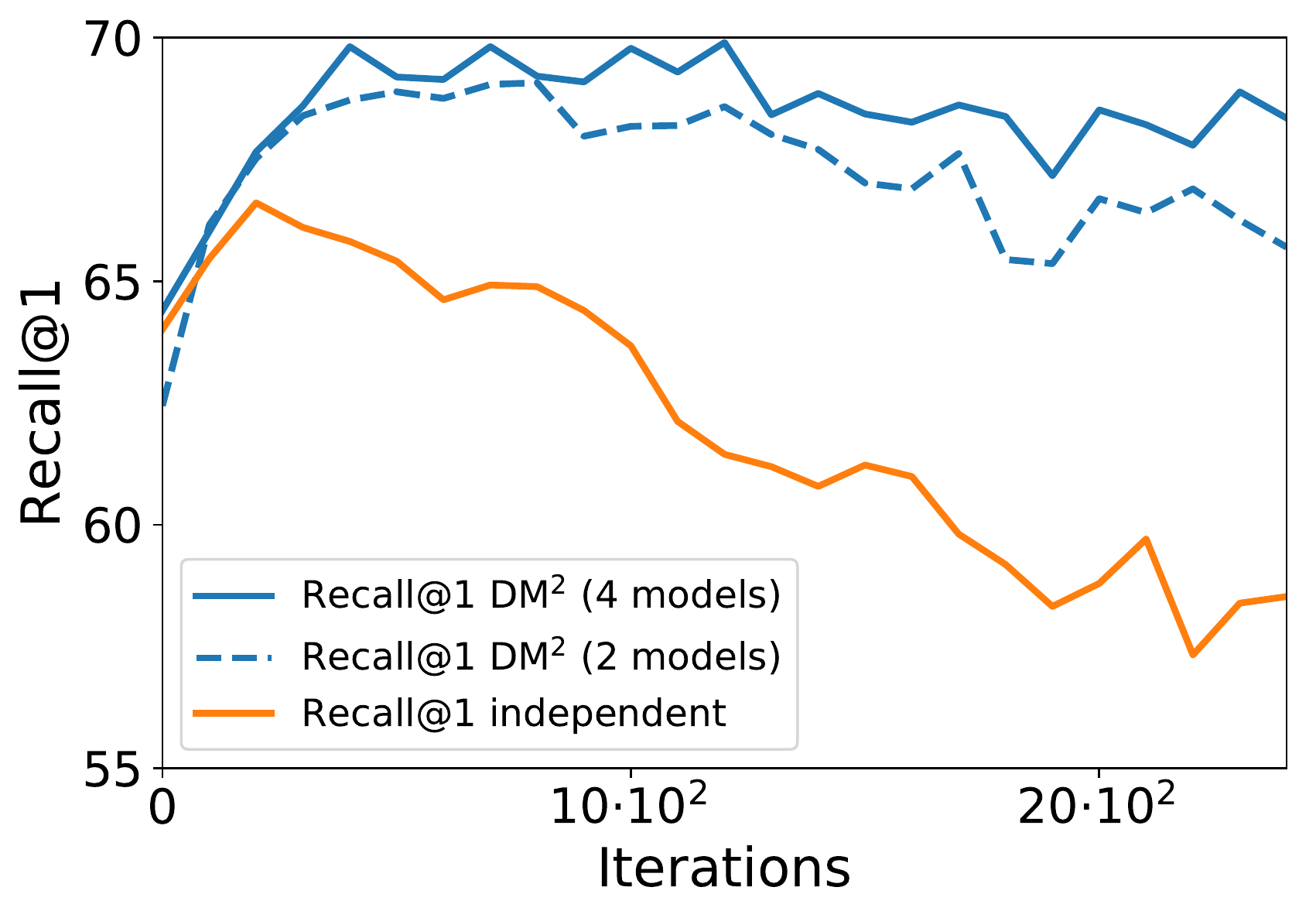}
        \caption{Test Recall@1 on CUB-200-2011}
    \end{subfigure}
    \begin{subfigure}{0.47\linewidth}
        \label{fig:reg_cars196}
        \includegraphics[width=\textwidth,height=4cm]{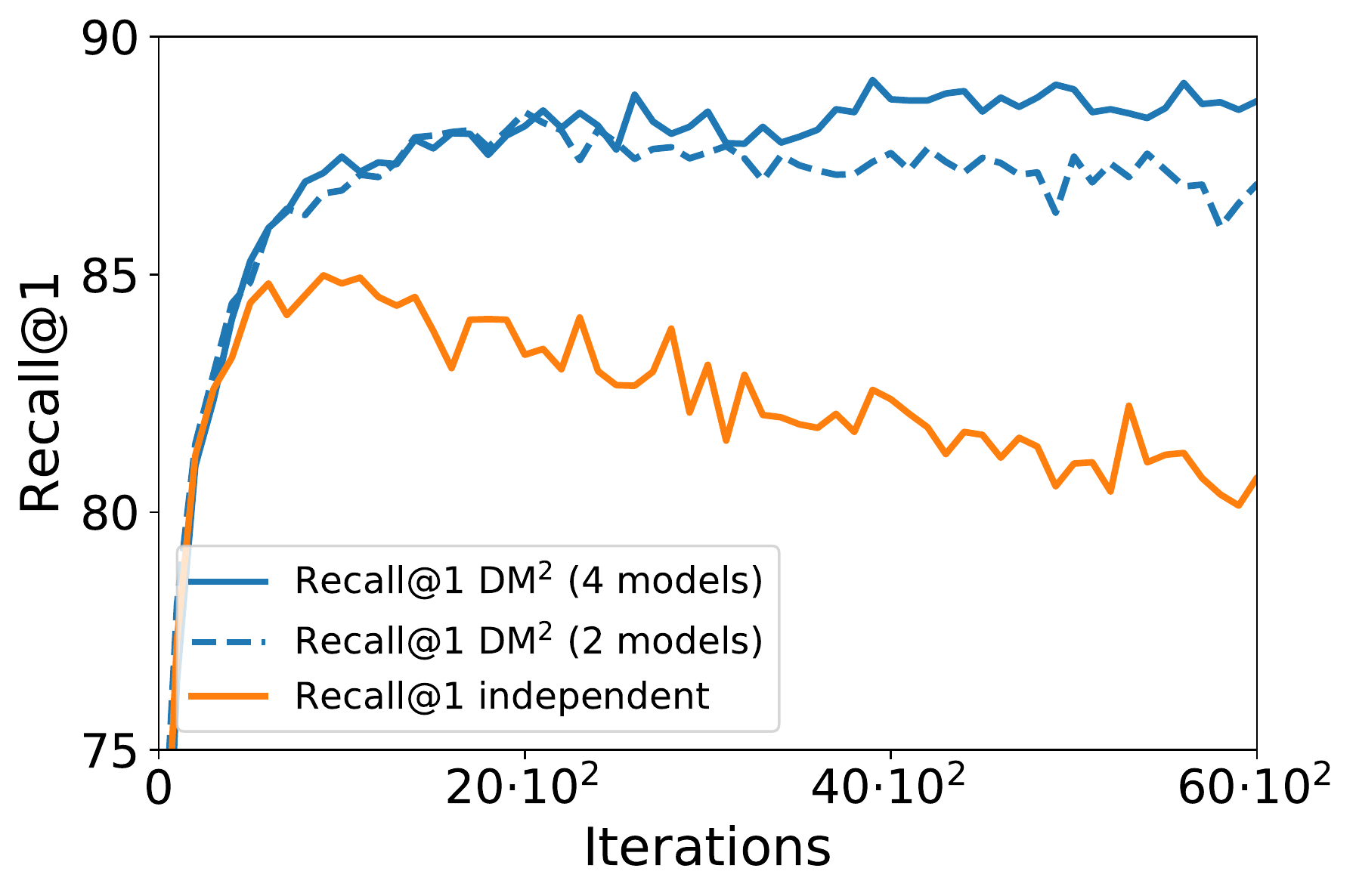}
        \caption{Test Recall@1 on Cars-196}
    \end{subfigure}
    \caption{Plots showing the changes of Recall@1 on the evaluation sets of CUB-200-2011 and Cars-196 during the training. For $\rmm$, the Recall@1 results are from the first model  ($l$=$1$) among the multiple models of the cohort.}
    \label{fig:regularizer}
\end{figure}

\subsection{Scalability of $\rmm$} \label{sec:scalability}

\Cref{fig:scale} delivers how the performance increases for larger cohort size.
For both datasets, CUB-200-2011 and Cars-196, the mutual learning with all three diversities (\ie, $\rmm$) results in a monotonic increment of Recall@1 along with the cohort size.
On the contrary, the ensemble performance of individually trained models converges faster at the lower Recall@1 than that of $\rmm$.

\Cref{fig:regularizer} depicts how mutual learning effects as a regularizer, which makes the model more robust.
The plot shows individually trained models suffer a problem of severe overfitting on the latter part of the training.
However, the model trained with $\rmm$ keeps outperforming the baseline without severe overfitting.
Note that its robustness keeps increase for a larger cohort size.

\begin{table*}[t]
    \centering
    \setlength{\tabcolsep}{1pt}
    \caption{Recall@K comparison with state-of-the-art metric learning methods.
    We divide methods with same backbone network. ``Dim.'' refers to the embedding dimension of the model. Boldface refers to the highest performance among same backbone. 
    The numbers under the datasets refer to recall at K. 
    The results of $\rmm$ are from a \textit{single} model for fair comparison, which is the first model ($l$$=$$1$) among multiple models of cohort.
    \\
    }
    \label{tab:sota_metriclearning}
    \label{table:1a}
    \begin{tabular}{|c|c|c|cccc|cccc|cccc|}
    \hline
     \multirow{2}{*}{\textbf{B}} & \multirow{2}{*}{Methods} &  \multirow{2}{*}{Dim.} & \multicolumn{4}{c|}{CUB-200-2011} & \multicolumn{4}{c|}{Cars-196} & \multicolumn{4}{c|}{SOP}\\
      &  & & 1 & 2 & 4 & 8 & 1 & 2 & 4 & 8 & 1 & 10 & $10^2$ & $10^3$  \\
    \hline
     \hline
     \multirow{5}{*}{\rotatebox[origin=c]{90}{ResNet-50}} & Margin~\cite{wu2017sampling} & 128 & 63.6 & 74.4 & 83.1 & 90.0 & 79.6 & 86.5 & 91.9 & 95.1 & 72.7 & 86.2 & 93.8 & 98.0\\
      & DC~\cite{sanakoyeu2019divide} & 128 & 65.9 & 76.6 & 84.4 & \textbf{90.6} & 84.6 & 90.7 & 94.1 & 96.5 & 75.9 & 88.4 & 94.9 & 98.1  \\ 
      & MIC~\cite{Roth_2019_ICCV} & 128 & 66.1 & 76.8 & \textbf{85.6} & - & 82.6 & 89.1 & 93.2 & - & 77.2 & 89.4 & 95.6 & - \\
      & TML~\cite{Yu_2019_ICCV} & 512 & 62.5 & 73.9 & 83.0 & 89.4 & 86.3 & 92.3 & \textbf{95.4} & \textbf{97.3} & 78.0 & 91.2 & 96.7 & 99.0 \\
      & Triplet + $\rmm$ & 128 & 64.1 & 75.0 & 83.2 & 89.4 & 84.7 & 89.9 & 93.5 & 95.8 & 79.9 & 91.3 & 96.6 & 99.0 \\  
      & Triplet + $\rmm$ & 512 & \textbf{66.7} & \textbf{77.1} & 85.2 & 90.5 & \textbf{86.9} & \textbf{92.0} & 94.5 & 96.8 & \textbf{80.4} & \textbf{91.8} & \textbf{96.8} & \textbf{99.1} \\
      \hline
      \hline
        
      \multirow{6}{*}{\rotatebox[origin=c]{90}{BN-Inception}} & HTL~\cite{ge2018deep} & 512 & 57.1 & 68.8 & 78.7 & 86.5 & 81.4 & 88.0 & 92.7 & 95.7 & 74.8 & 88.3 & 94.8 & 98.4\\
      & NSM~\cite{zhai2018making} & 512 & 59.6 & 72.0 & 81.2 & 88.4 & 81.7 & 88.9 & 93.4 & 96.0 & 73.8 & 88.1 & 95.0 & -\\
      & MS~\cite{wang2019multi} & 512 & 65.7 & 77.0 & 86.3 & 91.2 & 84.1 & 90.4 & 94.0 & 96.5 & 78.2 & 90.5 & 96.0 & 98.7\\
      & SoftTriplet~\cite{qian2019softtriple} & 512 & 65.4 & 76.5 & 84.5 & 90.4 & 84.5 & 90.7 & 94.5 & 96.9 & 78.3 & 90.3 & 95.9 & -\\
      & HORDE~\cite{JACOB_2019_ICCV} & 512 & 66.3 & 76.7 & 84.7 & 90.6 & 83.9 & 90.3 & 94.1 & 96.3 & \textbf{80.1} & \textbf{91.3} & 96.2 & 98.7 \\
      & Triplet + $\rmm$ & 512 & \textbf{69.9} & \textbf{79.7} & \textbf{86.5} & \textbf{91.4} & \textbf{89.1} & \textbf{93.3} & \textbf{95.8} & \textbf{97.6} & 78.8 & 90.9 & \textbf{96.3} & \textbf{98.9} \\
      \hline
    \end{tabular}   
\end{table*}
\begin{table}[!ht]
    \centering
    \setlength{\tabcolsep}{3pt}
    \caption{Recall@K comparison with state-of-the-art metric learning methods on In-Shop. “Dim.” refers to the embedding dimension of the model. Bold-face refers the highest performance among all models. 
    The results of $\rmm$ are from a \textit{single} model for fair comparison, which is the first model ($l$$=$$1$) among multiple models of cohort.\\}
    \label{tab:inshop_rkd}
  \begin{tabular}{|c|c|cccc|}
    \hline
    \multirow{2}{*}{Methods} & \multirow{2}{*}{Dim.} & \multicolumn{4}{c|}{In-Shop} \\
       &   & 1 & 10 & 20 & 30 \\
       \hline
          \hline

     FashionNet~\cite{liu2016deepfashion} & 4096 & 53.0 & 73.0 & 76.0 & 77.0\\
     NSM~\cite{zhai2018making} & 512 & 88.6 & 97.5 & 98.4 & \textbf{98.8}\\
     A-BIER~\cite{opitz_2018_pami} & 512 & 83.1 & 95.1 & 96.9 & 97.5 \\
     ABE-8\cite{Kim_2018_ECCV} & 512 & 87.3 & 96.7 & 97.9 & 98.2  \\
     MS~\cite{wang2019multi} & 512 & 89.7 & \textbf{97.9 }& 98.5 & \textbf{98.8}\\
     HORDE~\cite{JACOB_2019_ICCV} & 512 & \textbf{90.4} & 97.8 & 98.4 & 98.7 \\
     Triplet + $\rmm$ & 512 & 89.6 &  97.8 & \textbf{98.6} & \textbf{98.8} \\
    \hline
  \end{tabular}
\end{table}
\subsection{Comparison with Other Deep Metric Learning Studies}

Since the single model performance improved significantly with $\rmm$, we compared the recall performance of $\rmm$ with the current state-of-the-art deep metric learning methods.
For fair comparison, the number of embedding dimension and the backbone network were set according to those of compared methods.
The cohort size of $\rmm$ is fixed to $4$.

The result of \cref{tab:sota_metriclearning} shows the recalls measured on CUB-200-2011, Cars-196, and SOP.
For ResNet-50, the model of 512 embedding dimension trained with $\rmm$ and Triplet loss achieves the best Recall@1 for all three datasets: CUB-200-2011, Cars-196, and SOP.
For BN-Inception, it achieved the best Recall@1 for CUB-200-2011, Cars-196, and the second best Recall@1 for SOP.
For the results on In-shop dataset shown in \cref{tab:inshop_rkd}, our method achieves comparable performance with the best, less than 1\% point difference.


\begin{table}[t]
    \centering
    \caption{Recall@1 comparison between knowledge distillation methods for metric learning and $\rmm$. 
    $\textbf{P}$ indicates whether the teacher and student are trained in a parallel manner.
    For knowledge distillation methods, $\textbf{T}$ and $\textbf{S}$ refer to a teacher and a student respectively.
    For $\rmm$, $\textbf{T}$ and $\textbf{S}$ refer to an ensemble and a single model (the first model where $l$$=$$1$) respectively.
    `$\rightarrow$' means that two methods were applied in a sequence.
    \\}
        \setlength{\tabcolsep}{3pt}
        \begin{tabular}{|c|c|cccc|cccc|}
        \hline
         \multirow{3}{*}{Methods} & \multirow{3}{*}{\textbf{P}}  & \multicolumn{4}{c|}{CUB-200-2011} 
           & \multicolumn{4}{c|}{Cars-196} \\
          & & \multicolumn{2}{c}{2$\times$models} & \multicolumn{2}{c|}{4$\times$models} & \multicolumn{2}{c}{2$\times$models} & \multicolumn{2}{c|}{4$\times$models} \\
          & &  \textbf{T} & \textbf{S} &  \textbf{T} & \textbf{S} &  \textbf{T} & \textbf{S} &  \textbf{T} & \textbf{S} \\
          \hline
          \hline
          Baseline & X &  - & 66.61 & -  & 66.61 & - & 84.98 & - & 84.98 \\
          RKD-DA~\cite{park2019relational} & X & 69.36 & 70.13  & 70.68 & 70.64 & 88.26 & 88.82 & 89.59  & 89.95 \\
          DarkRank~\cite{chen2018darkrank} & X & 69.36 & 68.28  & 70.68 & 68.09  & 88.26 & 87.92 & 89.59 &  87.99 \\
            Compact~\cite{yu2019learning} & X & 69.36 & 68.70  & 70.68 & 67.98 & 88.26 & \textbf{89.50} & 89.59  & 89.48 \\        
          \hline
          \hline
         $\rmm$ & O & \textbf{70.68} & 69.07 & \textbf{72.89} & 69.90 & \textbf{89.57} & 88.41 & \textbf{91.48} & 89.10 \\
         $\rmm\rightarrow$RKD-DA & X & \textbf{70.68} & \textbf{70.54} & \textbf{72.89} & \textbf{71.22} & \textbf{89.57} & 89.14 & \textbf{91.48} & \textbf{90.21} \\
         \hline
        \end{tabular}
        \label{tab:rks_compare}
    \end{table}
    \smallbreak
\subsection{Combining $\rmm$ with Knowledge Distillation}

We combine the ensembled model of $\rmm$ with \textit{relational knowledge distillation} \cite{park2019relational}.
For a detailed comparison, we also report the results of other recently proposed knowledge distillation methods for deep metric learning \cite{chen2018darkrank,yu2019learning}.
To build a strong teacher model for the distillation methods, we build an ensemble of 4 independently trained models.
For $\rmm$, the cohort of size 4 is mutually trained in parallel. 
Each model is based on BN-Inception with the embedding dimension of 512, and we transfer the knowledge of the ensemble to a single model (student).
We apply the methods on the final representations, \ie, embeddings.
We describe each method briefly as follows:

\begin{itemize}
    \item \textbf{Relational Knowledge Distillation} \cite{park2019relational} transfers relative distance and angle formed by points in the embedding space of the model.
    Between the three combinations of loss they proposed, we used distance and angle combination (RKD-DA).
    Following the original paper, we did not apply $\mathcal{L}_{\text{DML}}$ while distillation and set the hyper-parameters $\lambda_{\text{distance}}$ and $\lambda_{\text{angle}}$ to 20 and 40 respectively.
    \item \textbf{DarkRank} \cite{chen2018darkrank} transfers similarity ranks between examples.
    Between the two losses they proposed, we use HardRank loss.
    We carefully tune the hyper-parameters for DarkRnak using a grid search.
    The hyper-parameters $\alpha$, $\beta$, and $\lambda_{\text{DarkRank}}$ are set to 3, 3, and 1, respectively.
    \item \textbf{Compact Networks} \cite{yu2019learning} transfers absolute distance among data examples.
    Following the original paper, the hyper-parameter $\lambda_{\text{compact}}$ is set to 10.
\end{itemize}

As shown in \cref{tab:rks_compare}, the result of pure $\rmm$ is on par with other two-staged distilled performances.
By combining $\rmm$ with the RKD-DA, which showed the best distillation performance, the final single model performance reaches the highest Recall@1 71.22 and 90.21 for CUB-200-2011 and Cars-196, respectively.
These results show that the ensemble model of $\rmm$ indeed holds better knowledge to transfer to the student (single model).

\begin{table}[t]
    \centering
    \caption{Recall@1 comparison varying the combinations of heterogenous cohorts. 
    BNI refers to BN-Inception and R50 refers to ResNet-50.  $\times$ indicates the number of cohorts with the architecture.
     The results of the Single are from the first model ($l$$=$$1$) among multiple models of cohort. Baseline is a single model trained independently without mutual learning with cohorts.\\}
        \setlength{\tabcolsep}{2pt}
        \begin{tabular}{|c|ccc|ccc|}
        \hline
        \multirow{2}{*}{Combinations} & \multicolumn{3}{c|}{CUB-200-2011} & \multicolumn{3}{c|}{Cars-196} \\
         & BNI (Single) & R50 (Single) & Ens & BNI (Single) & R50 (Single) & Ens \\
        \hline
        \hline
        Baseline & 66.61 & 64.01 & -  & 84.98 & 82.70 & - \\
        \hline
        \hline
        4$\times$BNI & \textbf{69.90} & - & 72.89 & \textbf{89.10} & - & 91.48 \\
        3$\times$BNI \; 1$\times$R50 & 68.69 & 66.63 & \textbf{74.17} & 87.13 & \textbf{88.76} & \textbf{91.76} \\
        2$\times$BNI \; 2$\times$R50 & 68.80 & \textbf{66.90} & 73.90 & 86.55 & 87.62 & 91.46 \\
        1$\times$BNI \; 3$\times$R50 & 68.08 & 66.44 & 71.59 & 86.41 & 86.20 & 90.21 \\
        4$\times$R50 & - & 66.71 & 68.87 & - & 86.93 & 88.62 \\
        

        \hline
        \end{tabular}
        \label{tab:heterogenous}
    \end{table}
    \smallbreak

\subsection{Heterogenous Cohort}

For this experiment, we conducted experiments on heterogeneous cohort, which comprises models with diverse architecture.
For that, we examined various construction for a cohort of size 4 using BN-Inception and ResNet-50, and the embedding dimension was 512.

\Cref{tab:heterogenous} shows the result of the experiment, and the row 4$\times$\text{BNI} is identical to our previous experimental setting.
We set the temporal diversity individually to the network type, for example, 3$\times$\text{BNI} 1$\times$\text{R50} setting has the update rate of $(1, 0.5, 0.25, 1)$ for three BNI and one R50, respectively.
The results show a feeble possibility of architecture diversity, and it is noteworthy that the  ensemble performances of CUB-200-2011 and Cars-196 were found to be better for the heterogeneous cohort than the homogenous cohort.
Also, one can see that the single model performance of R50 grows with the addition of BNI to its cohort.

\begin{table}[!ht]
    \centering
    \caption{Applicability of $\rmm$ on different deep metric learning objectives. 
    `Independent' refers to the case when a model is trained independently with only $\mathcal{L}_\text{DML}$.
    For $\rmm$, the results of Single Model are from the first model ($l$$=$$1$) among multiple models of cohort.\\}
    \label{tab:applicable}
        \setlength{\tabcolsep}{3pt}
        \begin{subtable}{\linewidth}\centering
        \caption{Recall@1 on CUB-200-2011.}
        \begin{tabular}{|c|cc|cc|}
        \hline
         \multirow{2}{*}{$\mathcal{L}_\text{DML}$} & \multicolumn{2}{c|}{Independent} & \multicolumn{2}{c|}{$\rmm$}\\
          & Single Model & Ensemble & Single Model & Ensemble \\
         \hline
         \hline
         Contrastive~\cite{hadsell2006dimensionality} & 61.97 & 63.94 & 66.40  & 67.45 \\
         Triplet~\cite{DBLP:journals/corr/SchroffKP15} & 66.07 & 70.71 & 69.90 & 72.89 \\
         Binomial-Deviance~\cite{yideep2014} & 64.83 & 69.01 & 67.92 & 69.90 \\
         Proxy-NCA~\cite{movshovitz2017proxynca} & 60.11 & 67.42 & 64.95 & 68.67 \\
         ArcFace~\cite{deng2019arcface}  & 62.09 & 69.05 & 67.74 & 70.32 \\
         \hline
        \end{tabular}
        \end{subtable}
        
        \begin{subtable}{\linewidth}\centering
        \caption{Recall@1 on Cars-196.}
        \begin{tabular}{|c|cc|cc|}
        \hline
         \multirow{2}{*}{$\mathcal{L}_\text{DML}$} & \multicolumn{2}{c|}{Independent} & \multicolumn{2}{c|}{$\rmm$}\\
         & Single Model & Ensemble & Single Model & Ensemble \\
         \hline
         \hline
         Contrastive~\cite{hadsell2006dimensionality} & 78.93 & 82.97 & 83.10 & 86.72  \\
         Triplet~\cite{DBLP:journals/corr/SchroffKP15} & 85.24 & 89.87 & 89.10  & 91.48 \\
         Binomial-Deviance~\cite{yideep2014} & 83.94 & 89.16 & 88.33 & 89.20  \\
         Proxy-NCA~\cite{movshovitz2017proxynca} & 83.70  & 90.94  & 86.74  & 89.01  \\
         ArcFace~\cite{deng2019arcface}  & 85.11 & 91.13 & 88.45 & 90.16 \\
         \hline
        \end{tabular}
        \end{subtable}
    \end{table}

\subsection{$\rmm$ with Other $\mathcal{L}_{\text{DML}}$} \label{exp:different_losses}

$\rmm$ can be effective with any deep metric learning objectives.
To validate that, we conduct an experiment of combining $\rmm$ with widely adopted deep metric learning losses, \eg, Triplet, Contrastive, Proxy-NCA, Binomial-Deviance, and ArcFace.
The size of the cohort is set to 4, the models of BN-Inception with 512 embedding dimension are adopted.
We describe the hyper-parameters of each method.

\begin{itemize}
    \item \textbf{Triplet \cite{DBLP:journals/corr/SchroffKP15}}: we follow the hyper parameter specified in \cref{sec:impl_detail}.
    \item \textbf{Contrastive \cite{hadsell2006dimensionality}}: we set the margin to 1.0 and $\lambda_{\rmm}$ to 20. 
    \item \textbf{Proxy-NCA \cite{movshovitz2017proxynca}}: we set the number of the proxies to 1 and $\lambda_{\rmm}$ to $10^2$.
    \item \textbf{Binomial-Deviance \cite{yideep2014}}: we set the alpha and beta to 2 and 0.5 respectively following \cite{yideep2014}, and $\lambda_{\rmm}$ to 20.
    \item \textbf{ArcFace \cite{deng2019arcface}}: we set the scale factor and margin to 64 and 0.5 respectively following \cite{deng2019arcface}, and $\lambda_{\rmm}$ to $2\cdot 10^{3}$.
\end{itemize}

The result in \cref{tab:applicable} delivers that $\rmm$ improves the performance of a \textit{single model} with any deep metric learning objectives.

\section{Conclusion} \label{sec:conclusion}
In this paper, we propose \textbf{D}iversified \textbf{M}utual \textbf{M}etric Learning ($\rmm$) to effectively apply mutual learning on the task of deep metric learning.
In $\rmm$, three diversities are proposed to enrich the knowledge shared by the cohort of mutual learning: model diversity, temporal diversity, and view diversity. 
We proved that the temporal and view diversities were especially essential to the application of mutual learning since deep metric learning requires pretrained models due to its small size dataset.
By combining all three diversities carefully, $\rmm$ results the state-of-the-art performance on widely adopted deep metric learning datasets: CUB-200-2011 and Cars-196, and SOP.


\section*{Acknowledgement}
This work was partly supported by Institute for Information \& communications Technology Promotion (IITP) grant funded by the Korea government (MSIP) (No. 2019-0-01906, Artificial Intelligence Graduate School Program (POSTECH)) and  Basic Science Research Program (NRF-2017R1E1A1A01077999) through the National Research Foundation of Korea (NRF) funded by the Ministry of Science, ICT.

%
%
\bibliographystyle{splncs04}
\bibliography{main}

\begin{thebibliography}{10}
\providecommand{\url}[1]{\texttt{#1}}
\providecommand{\urlprefix}{URL }
\providecommand{\doi}[1]{https://doi.org/#1}

\bibitem{anil2018large}
Anil, R., Pereyra, G., Passos, A., Ormandi, R., Dahl, G.E., Hinton, G.E.: Large
  scale distributed neural network training through online distillation. arXiv
  preprint arXiv:1804.03235  (2018)

\bibitem{chen2018darkrank}
Chen, Y., Wang, N., Zhang, Z.: Darkrank: Accelerating deep metric learning via
  cross sample similarities transfer. In: Thirty-Second AAAI Conference on
  Artificial Intelligence (2018)

\bibitem{deng2019arcface}
Deng, J., Guo, J., Xue, N., Zafeiriou, S.: Arcface: Additive angular margin
  loss for deep face recognition. In: Proceedings of the IEEE Conference on
  Computer Vision and Pattern Recognition. pp. 4690--4699 (2019)

\bibitem{ge2018deep}
Ge, W.: Deep metric learning with hierarchical triplet loss. In: Proceedings of
  the European Conference on Computer Vision (ECCV). pp. 269--285 (2018)

\bibitem{hadsell2006dimensionality}
Hadsell, R., Chopra, S., LeCun, Y.: Dimensionality reduction by learning an
  invariant mapping. In: 2006 IEEE Computer Society Conference on Computer
  Vision and Pattern Recognition (CVPR'06). vol.~2, pp. 1735--1742. IEEE (2006)

\bibitem{DBLP:journals/corr/HeZRS15}
He, K., Zhang, X., Ren, S., Sun, J.: Deep residual learning for image
  recognition. In: The IEEE Conference on Computer Vision and Pattern
  Recognition (CVPR) (2016)

\bibitem{hinton2015distilling}
Hinton, G., Vinyals, O., Dean, J.: Distilling the knowledge in a neural
  network. arXiv preprint arXiv:1503.02531  (2015)

\bibitem{huang2017like}
Huang, Z., Wang, N.: Like what you like: Knowledge distill via neuron
  selectivity transfer. arXiv preprint arXiv:1707.01219  (2017)

\bibitem{ioffe2015batch}
Ioffe, S., Szegedy, C.: Batch normalization: Accelerating deep network training
  by reducing internal covariate shift. arXiv preprint arXiv:1502.03167  (2015)

\bibitem{JACOB_2019_ICCV}
Jacob, P., Picard, D., Histace, A., Klein, E.: Metric learning with horde:
  High-order regularizer for deep embeddings. In: The IEEE International
  Conference on Computer Vision (ICCV) (Oct 2019)

\bibitem{Kim_2018_ECCV}
Kim, W., Goyal, B., Chawla, K., Lee, J., Kwon, K.: Attention-based ensemble for
  deep metric learning. In: The European Conference on Computer Vision (ECCV)
  (2018)

\bibitem{kingma2014adam}
Kingma, D.P., Ba, J.: Adam: A method for stochastic optimization. arXiv
  preprint arXiv:1412.6980  (2014)

\bibitem{KrauseStarkDengFei-Fei_3DRR2013}
Krause, J., Stark, M., Deng, J., Fei-Fei, L.: 3d object representations for
  fine-grained categorization. In: 4th International IEEE Workshop on 3D
  Representation and Recognition (3dRR-13) (2013)

\bibitem{liu2019knowledge}
Liu, Y., Cao, J., Li, B., Yuan, C., Hu, W., Li, Y., Duan, Y.: Knowledge
  distillation via instance relationship graph. In: Proceedings of the IEEE
  Conference on Computer Vision and Pattern Recognition. pp. 7096--7104 (2019)

\bibitem{liu2016deepfashion}
Liu, Z., Luo, P., Qiu, S., Wang, X., Tang, X.: Deepfashion: Powering robust
  clothes recognition and retrieval with rich annotations. In: Proceedings of
  the IEEE conference on computer vision and pattern recognition. pp.
  1096--1104 (2016)

\bibitem{movshovitz2017proxynca}
Movshovitz{-}Attias, Y., Toshev, A., Leung, T.K., Ioffe, S., Singh, S.: No fuss
  distance metric learning using proxies. CoRR  \textbf{abs/1703.07464} (2017),
  \url{http://arxiv.org/abs/1703.07464}

\bibitem{Song2016DeepML}
Oh~Song, H., Xiang, Y., Jegelka, S., Savarese, S.: Deep metric learning via
  lifted structured feature embedding. In: The IEEE Conference on Computer
  Vision and Pattern Recognition (CVPR) (2016)

\bibitem{opitz_2018_pami}
Opitz, M., Waltner, G., Possegger, H., Bischof, H.: Deep metric learning with
  bier: Boosting independent embeddings robustly. IEEE Transactions on Pattern
  Analysis and Machine Intelligence  (2018). \doi{10.1109/TPAMI.2018.2848925}

\bibitem{park2019relational}
Park, W., Kim, D., Lu, Y., Cho, M.: Relational knowledge distillation. In:
  Proceedings of the IEEE Conference on Computer Vision and Pattern
  Recognition. pp. 3967--3976 (2019)

\bibitem{paszke2017automatic}
Paszke, A., Gross, S., Chintala, S., Chanan, G., Yang, E., DeVito, Z., Lin, Z.,
  Desmaison, A., Antiga, L., Lerer, A.: Automatic differentiation in pytorch
  (2017)

\bibitem{qian2019softtriple}
Qian, Q., Shang, L., Sun, B., Hu, J., Li, H., Jin, R.: Softtriple loss: Deep
  metric learning without triplet sampling. arXiv preprint arXiv:1909.05235
  (2019)

\bibitem{romero2014fitnets}
Romero, A., Ballas, N., Kahou, S.E., Chassang, A., Gatta, C., Bengio, Y.:
  Fitnets: Hints for thin deep nets. International Conference on Learning
  Representations  (2015)

\bibitem{Roth_2019_ICCV}
Roth, K., Brattoli, B., Ommer, B.: Mic: Mining interclass characteristics for
  improved metric learning. In: The IEEE International Conference on Computer
  Vision (ICCV) (October 2019)

\bibitem{ILSVRC15}
Russakovsky, O., Deng, J., Su, H., Krause, J., Satheesh, S., Ma, S., Huang, Z.,
  Karpathy, A., Khosla, A., Bernstein, M., Berg, A.C., Fei-Fei, L.: {ImageNet
  Large Scale Visual Recognition Challenge}. International Journal of Computer
  Vision (IJCV)  \textbf{115}(3),  211--252 (2015).
  \doi{10.1007/s11263-015-0816-y}

\bibitem{sanakoyeu2019divide}
Sanakoyeu, A., Tschernezki, V., Buchler, U., Ommer, B.: Divide and conquer the
  embedding space for metric learning. In: Proceedings of the IEEE Conference
  on Computer Vision and Pattern Recognition. pp. 471--480 (2019)

\bibitem{DBLP:journals/corr/SchroffKP15}
Schroff, F., Kalenichenko, D., Philbin, J.: Facenet: A unified embedding for
  face recognition and clustering. In: The IEEE Conference on Computer Vision
  and Pattern Recognition (CVPR) (2015)

\bibitem{suh2019stochastic}
Suh, Y., Han, B., Kim, W., Lee, K.M.: Stochastic class-based hard example
  mining for deep metric learning. In: Proceedings of the IEEE Conference on
  Computer Vision and Pattern Recognition. pp. 7251--7259 (2019)

\bibitem{ustinova2016learning}
Ustinova, E., Lempitsky, V.: Learning deep embeddings with histogram loss. In:
  Advances in Neural Information Processing Systems. pp. 4170--4178 (2016)

\bibitem{WahCUB_200_2011}
Wah, C., Branson, S., Welinder, P., Perona, P., Belongie, S.: {The Caltech-UCSD
  Birds-200-2011 Dataset}. Tech. Rep. CNS-TR-2011-001, California Institute of
  Technology (2011)

\bibitem{wang2019multi}
Wang, X., Han, X., Huang, W., Dong, D., Scott, M.R.: Multi-similarity loss with
  general pair weighting for deep metric learning. In: Proceedings of the IEEE
  Conference on Computer Vision and Pattern Recognition. pp. 5022--5030 (2019)

\bibitem{wu2017sampling}
Wu, C.Y., Manmatha, R., Smola, A.J., Krahenbuhl, P.: Sampling matters in deep
  embedding learning. In: The IEEE International Conference on Computer Vision
  (ICCV) (2017)

\bibitem{yideep2014}
Yi, D., Lei, Z., Li, S.: Deep metric learning for practical person
  re-identification. ArXiv e-prints  (2014)

\bibitem{Yu_2019_ICCV}
Yu, B., Tao, D.: Deep metric learning with tuplet margin loss. In: The IEEE
  International Conference on Computer Vision (ICCV) (October 2019)

\bibitem{yu2019learning}
Yu, L., Yazici, V.O., Liu, X., Weijer, J.v.d., Cheng, Y., Ramisa, A.: Learning
  metrics from teachers: Compact networks for image embedding. In: Proceedings
  of the IEEE Conference on Computer Vision and Pattern Recognition. pp.
  2907--2916 (2019)

\bibitem{yuan2017hard}
Yuan, Y., Yang, K., Zhang, C.: Hard-aware deeply cascaded embedding. In:
  Proceedings of the IEEE international conference on computer vision. pp.
  814--823 (2017)

\bibitem{zagoruyko2016paying}
Zagoruyko, S., Komodakis, N.: Paying more attention to attention: Improving the
  performance of convolutional neural networks via attention transfer.
  International Conference on Learning Representations  (2017)

\bibitem{zhai2018making}
Zhai, A., Wu, H.Y.: Classification is a strong baseline for deep metric
  learning. arXiv preprint arXiv:1811.12649  (2018)

\bibitem{zhaiclassification2019}
Zhai, A., Wu, H.Y., San~Francisco, U.: Classification is a strong baseline for
  deep metric learning  (2019)

\bibitem{zhang2018deep}
Zhang, Y., Xiang, T., Hospedales, T.M., Lu, H.: Deep mutual learning. In:
  Proceedings of the IEEE Conference on Computer Vision and Pattern
  Recognition. pp. 4320--4328 (2018)

\end{thebibliography}

\end{document}